\documentclass[10pt, a4paper]{article}

\usepackage[]{lrec-coling2024} 
\usepackage{float}
\usepackage{todonotes}
\usepackage{amssymb}
\usepackage{array}
\usepackage{amsmath}
\usepackage{arydshln}
\usepackage{pifont}

\usepackage{graphicx}
\usepackage{tabularx}
\usepackage{xcolor}
\usepackage{svg}

\usepackage{nopageno}

\newcommand{\blfootnote}[1]{%
    \begingroup
    \renewcommand{\thefootnote}{}
    \footnote{#1}
    \addtocounter{footnote}{-1}
    \endgroup
}

\definecolor{light-gray}{gray}{0.95}
\newcommand{\code}[1]{\colorbox{light-gray}{\texttt{#1}}}

\newcommand{\cmark}{\ding{51}}%
\newcommand{\xmark}{\ding{55}}%

\graphicspath{{.}}

\title{The IgboAPI Dataset: Empowering Igbo Language Technologies through Multi-dialectal Enrichment}

\name{Chris Chinenye Emezue$^{*,2,3}$, Ifeoma Okoh$^{*,3}$, \\
Chinedu Mbonu$^{5}$, Chiamaka Chukwuneke$^{4}$,
    Daisy Lal$^{4}$, Ignatius Ezeani$^{4}$, Paul Rayson$^{4}$, Ijemma Onwuzulike$^{1}$,\\
    Chukwuma Okeke$^{1,6}$, Gerald Nweya$^{1}$, Bright Ogbonna$^{6}$, \\
    Chukwuebuka ORAEGBUNAM$^{1,7}$, Esther Chidinma Awo-Ndubuisi$^{6}$,\\
    Akudo Amarachukwu Osuagwu$^{7}$, Obioha Nmezi$^{1}$}
\vspace{0.5cm}
\address{$^{1}$Nk\d{o}wa okwu, $^{2}$Lanfrica,$^{3}$Masakhane, $^{4}$UCREL, Lancaster University, UK, \\ 
$^{5}$Nnamdi Azikiwe University, Nigeria, $^{6}$ University of Nigeria, Nsukka, $^{7}$University of Ibadan \\
          kedu@nkowaokwu.com,chris.emezue@lanfrica.com\\
         }

\abstract{
The Igbo language is facing a risk of becoming endangered, as indicated by a 2025 UNESCO study. This highlights the need to develop language technologies for Igbo to foster communication, learning and preservation. To create robust, impactful, and widely adopted language technologies for Igbo, it is essential to incorporate the multi-dialectal nature of the language. The primary obstacle in achieving dialectal-aware language technologies is the lack of comprehensive dialectal datasets. In response, we present the IgboAPI dataset, a multi-dialectal Igbo-English dictionary dataset, developed with the aim of enhancing the representation of Igbo dialects. Furthermore, we illustrate the practicality of the IgboAPI dataset through two distinct studies: one focusing on Igbo semantic lexicon and the other on machine translation. In the semantic lexicon project, we successfully establish an initial Igbo semantic lexicon for the Igbo semantic tagger, while in the machine translation study, we demonstrate that by finetuning existing machine translation systems using the IgboAPI dataset, we significantly improve their ability to handle dialectal variations in sentences.\\ \newline }

\begin{document}

\maketitleabstract

\section{Introduction}
\blfootnote{* Equal contribution}
The Igbo language is one of the three major languages in Nigeria, spoken by approximately 30 million people worldwide \cite{ethnologue}. Despite its significant population, Igbo culture is grappling with a form of social violence against its language in various Nigerian contexts \cite{article,unesco2}, as well as dwindling interest among the younger generation \cite{emekanwobia19language}. Igbo has been relegated to a secondary status when compared to English, which is widely perceived by many Nigerians as the language of prosperity and opportunity. This pervasive social issue has raised concerns to the extent that UNESCO has projected a risk of Igbo language becoming extinct by 2025 \cite{unesco2}. One key factor contributing to this perceived social violence against the Igbo language is the multi-dialectal nature of the language \cite{nwaozuzu2008dialects}, which has made it challenging for linguistic initiatives, lexical tools and language technologies that solely focus on the `Standard Igbo' to gain widespread acceptance, particularly among the broader language-speaking community. It is crucial for speakers of these dialects to feel included, but the multitude of dialects complicates efforts to accommodate them. 

In the past decade, we have witnessed incredible technological advancements surrounding language technologies and natural language processing. Language technology has been applied to solve many real-world problems that revolve around language \cite{hirschberg2015advances}. In today's globalized world, language technology plays a pivotal role in promoting, and preserving languages \cite{abbott2018towards,nekoto-etal-2020-participatory}. Resources like language courses, dictionaries, language learning apps, translation systems, education materials, audio resources, and language software can facilitate the documentation, teaching, and learning of the language in an easy way, serving as a platform for encouraging the younger generation to engage with their roots and heritage and ensuring the language thrives in contemporary settings \cite{oparaefefct2016,nwankwere2017safeguarding,anyanwu2019igbo}. Essentially, language technologies could benefit Igbo speakers, enrich global linguistic and cultural diversity, and prevent the extinction of the language.


We argue that embracing the rich diversity of Igbo dialects in the development of language technologies for the Igbo language is a fundamental step towards making such resources robust, effective and more widely accepted. The diversity of the Igbo language highlights a pressing need for linguistic tools tailored to the Igbo language and its varied dialects \cite{anyanwu2010igbo}. 

One major challenge that has hindered the progress of dialectal-aware language technologies lies in the scarcity of comprehensive datasets that represent these distinct Igbo dialects \cite{joshi-etal-2020-state}. Recognizing the critical need for such dialectal-aware resources, our IgboAPI project emerges as a pioneering effort to address this deficiency. By curating and making accessible a multi-dialectal dataset, we are not only enriching the linguistic landscape but also equipping language systems with the necessary tools to navigate the dialectal mosaic of the Igbo language. In this paper, we demonstrate the utility of the IgboAPI dataset in two applications. 

The remainder of this paper is organized as follows: in section \ref{relatedWork}, we begin by presenting relevant prior research, followed by an in-depth exploration of the IgboAPI project, in section \ref{methodology}, which encompasses the IgboAPI project, the dataset creation process and useful statistics about the resulting IgboAPI dictionary dataset. Moving on to section \ref{experiments}, we outline the two studies conducted to underscore the utility of the IgboAPI dictionary dataset. We describe the experiments in detail and end with results and valuable discussions.

\section{Related Work}
\label{relatedWork}
 The Igbo language falls under the `left-behinds' category, as classified by \citet{joshi-etal-2020-state}, meaning that it has received minimal attention in the realm of language technologies, and the availability of language technology datasets is notably scarce. Nevertheless, there have been increasing endeavors to create lexical resources \cite{gbal1962kwaOkwuI,green1971,Nnaji1985,Eke2001,igboanusiglossary,Mbah2021} and natural language processing datasets \cite{10.1145/3146387,ezeani2020igbo,adelani2022few} for the Igbo language. Notably, historically significant dictionary resources were pioneered by \citet{gbal1962kwaOkwuI} and \citet{Nnaji1985}. More recently, in the context of natural language processing, \citet{ezeani2020igbo} established a benchmark dataset comprising 5,630 English sentences that were translated into Igbo. Additionally, they translated 5,503 collected Igbo sentences into English through human intervention, resulting in English-Igbo sentence pairs. Another significant contribution is the JW300 dataset \cite{agic2019jw300} which provides a substantial corpus with a primary focus on the religious domain. Moreover, the IgboSum1500 \cite{mbonu2022igbosum1500} initiative has created an Igbo text summarization dataset, housing 1,500 articles. Our unique contribution lies in the inclusion of the various dialects present in the Igbo language, an aspect previously unexplored as all the aforementioned works dealt solely (or mostly) with the Standard Igbo.

The potential impact of dialectal diversity and its inclusion in the development of language technologies, such as lexicons and machine translation systems, has remained relatively unexplored within the Igbo language context. 
Some studies shed light on this for other languages. For instance, \citet{abe2018multi} explored multi-dialectal neural machine translation (NMT) from Japanese dialects to standard Japanese, emphasizing the potential benefits, particularly for an aging population more familiar with regional dialects. \citet{almansor2017translating} tackled translation from Egyptian Arabic dialect to Modern Standard Arabic. Leveraging our dialectal-aware dataset, we conduct experiments in Igbo-English translation, providing valuable insights into the impact of training a machine translation (MT) with a dialectal-aware dataset.



\section{The IgboAPI Project}
\label{methodology}


\subsection{Background}
The IgboAPI Project was created to address the pervasive problem experienced by many mostly new-generation Nigerians eager to learn the Igbo language: the lack of readily available, high-quality lexical and learning resources to support Igbo language learning. The IgboAPI Project focuses on addressing this problem by fully cataloging and annotating the linguistic nature of the Igbo language community contributions and feedback in the form of suggesting, adding, and reviewing dictionary data.



The IgboAPI project is structured in that it contains Igbo words and example sentences which can be fully annotated and interconnected with each other. Similar to WordNet \cite{wordnet}, the IgboAPI interconnects related words to each other and words to example sentences that feature specific usage of the word. Unlike the WordNet, though, the IgboAPI does not have a strong sense of linguistic hierarchy when connecting words with each other. This decision was made to prioritize collecting, arguably, more important pieces of word and example sentence metadata. The only lexical hierarchy that’s defined in the IgboAPI is the uni-direction connection of a word to its word stem.

\subsection{Creating the IgboAPI Dataset}
\begin{figure*}[!h]
\resizebox{\textwidth}{!}{
    \includegraphics[]{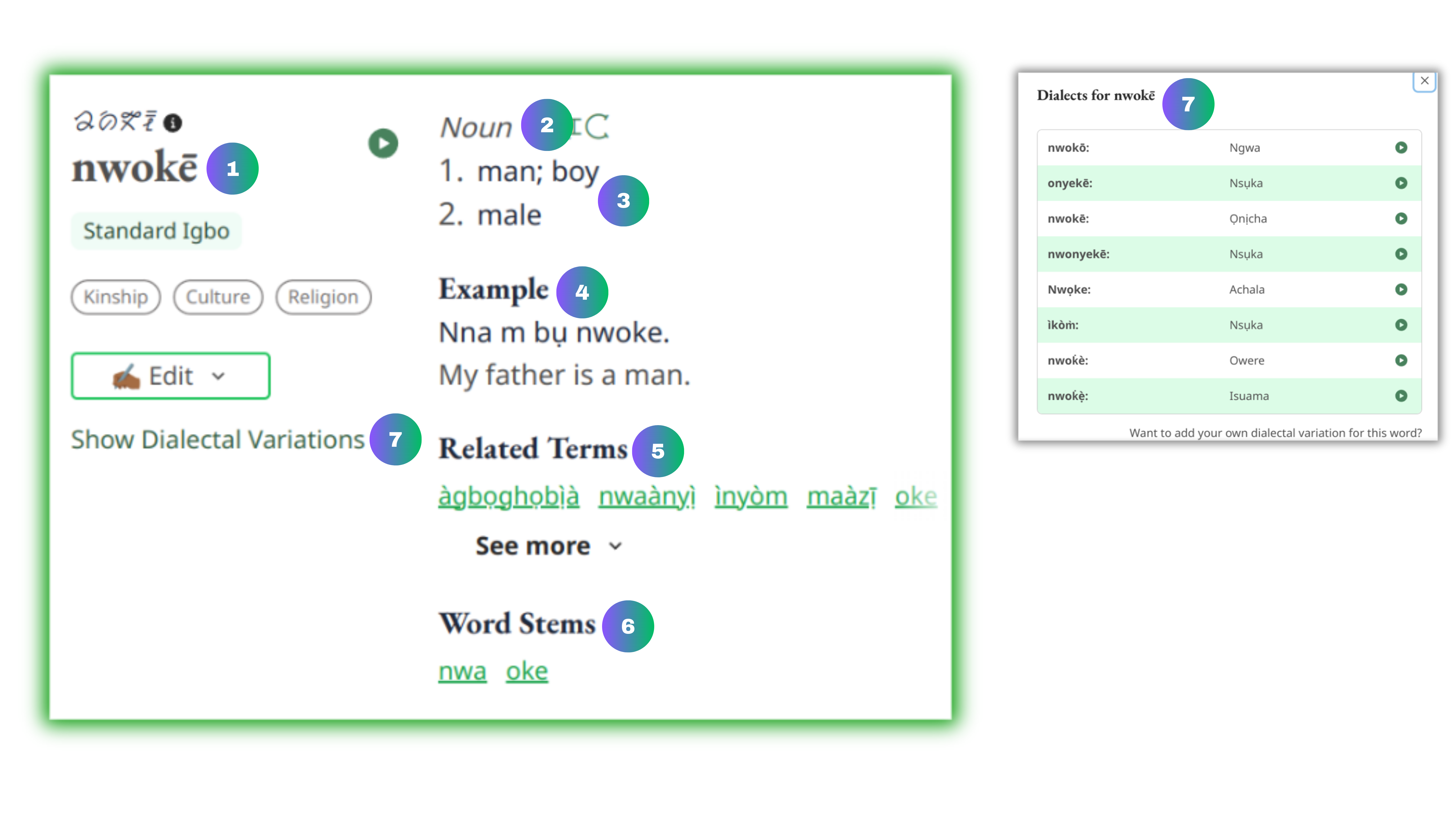}
    }
    
\caption{Illustrating the attributes of each word entry in the IgboAPI dataset using the example \textit{nwok\={e}}. Each sourced Igbo word contains the following attributes: 1) the headword which is the Standard Igbo variant of the word; 2) part of speech; 3) English definition of the Igbo word; 4) sentence of examples of the word (provided in both Igbo and English languages). The required number of example sentences is 1-4; 5) other words related to the given Igbo word; 6) word stems; 7) dialectal variations of the Igbo word. The dialectal variations provided depend on the lexicographers.}
\label{fig:igboapi-word}
\end{figure*}

The IgboAPI dataset is a multidialectal, Igbo-English dictionary dataset, created by expert lexicographers sourcing for Igbo words, and adding required metadata, including their dialectal variations. Each Igbo word entry required the attributes illustrated in Figure \ref{fig:igboapi-word}.

\paragraph{Key Participants and Roles:}
There were eight Igbo lexicographers, two Ns\d{i}b\d{i}d\d{i} lexicographers, two software engineers, one project manager, and three project owners, making up the total core team size of 16 members. ``Project owners'' were responsible for ensuring that the project was on track for the entire 12 months. They also led the team recruitment and training as well as engaged in regular meetings. The 
``Igbo Lexicographers'' had the task of sourcing for Igbo words and adding them, along with their dialectal variations and example sentences to the IgboAPI Editor platform. Lexicographers were also responsible for reviewing each other's work for quality assurance. The ``Ns\d{i}b\d{i}d\d{i} lexicographers'' mainly focused on adding Ns\d{i}b\d{i}d\d{i} script to all the words and example sentences added by the lexicographers. Finally, the ``Software Engineers'' were responsible for maintaining the IgboAPI Editor Platform by fixing bugs and implementing features to improve the work process for lexicographers.

\paragraph{Lexicography Tasks:} All tasks were performed using the IgboAPI Editor Platform and the dictionary editing standards designed by the IgboAPI project team \cite{dictEditingStandards}. Creating the IgboAPI dataset involved the lexicographers performing two major tasks: ``completing words'' and ``reviewing words''. ``Completing'' Igbo words involved fully annotating the sourced Igbo words with their required attributes. A fully annotated Igbo word would include all the required attributes (see Figure \ref{fig:igboapi-word}).


\paragraph{Quality Assurance:}
For quality assurance, we utilized the expertise of our lexicographers, designating them as ``reviewers'' responsible for evaluating the submissions of other lexicographers. This approach significantly streamlined the process and was time effective.
\paragraph{Training:}
All the lexicographers underwent training and onboarding before working. During training, lexicographers were responsible for reading through the Nk\d{o}wa okwu Dictionary Editing Standards documentation \footnote{\url{https://bit.ly/3FmXkH1}}, a comprehensive guide on their lexicography tasks. Training also included a presentation on how to identify data collection (sourcing for Igbo words in this case) bias that could lead to a lower-quality dataset. 

\subsection{The IgboAPI dataset}
\begin{figure*}[!h]
\begin{center}
\resizebox{0.6\textwidth}{!}{
\includegraphics{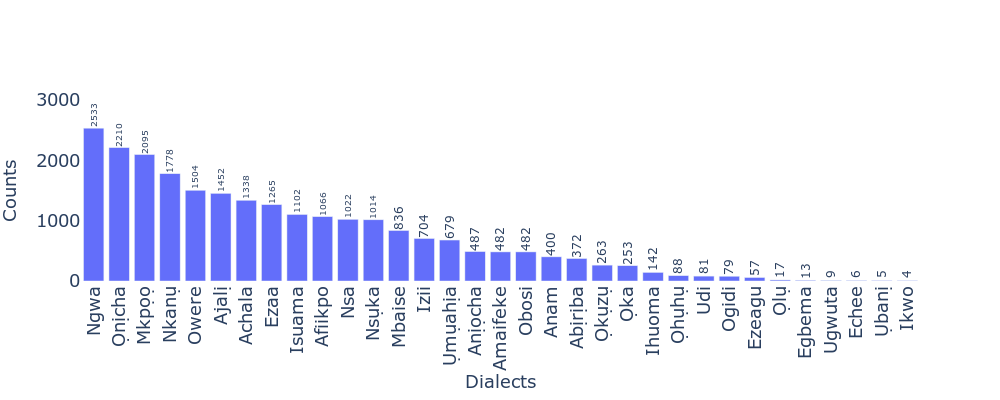} 
}
\caption{Dialects covered in the IgboAPI dataset and their number of words.}
\label{fig.1}
\end{center}
\end{figure*}
The IgboAPI dataset (also referred to as the IgboAPI dictionary dataset) is a multidialectal, multi-purpose Igbo-English dictionary dataset. It is multi-purpose in the sense that the various attributes of each word enables one to repurpose the dictionary dataset into other datasets for various NLP tasks: for example, one can use the Igbo-English example sentences as a parallel corpora for machine translation, or the audio-text pairs for speech processing, or the word classes information for part of speech tagging. Furthermore, the presence of dialectal variations for each word allows for easy substitution of Standard Igbo words with their dialectal equivalents, leading to a dataset that is inclusive of the Igbo dialects.

\begin{figure}[H]
\begin{center}
\resizebox{\columnwidth}{!}{
\includegraphics[]{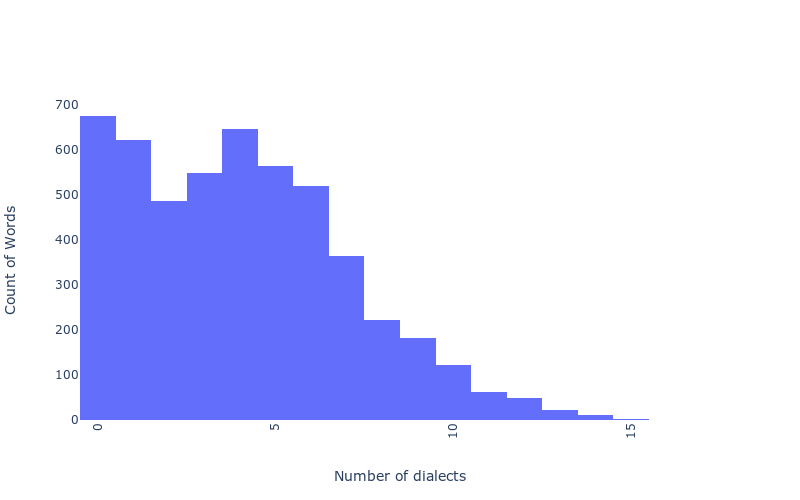}
}
\caption{Histogram of the count of dialects for each word. Each word can possess one or more Igbo dialects.  Many words in our dataset have more than one dialect (with some words having as high as 10 diverse dialectal variations. This highlights the multidialectal nature of the Igbo language.}
\label{fig.2}
\end{center}
\end{figure}

\begin{figure*}[!h]
\begin{center}
\resizebox{0.5\textwidth}{!}{
\includegraphics[]{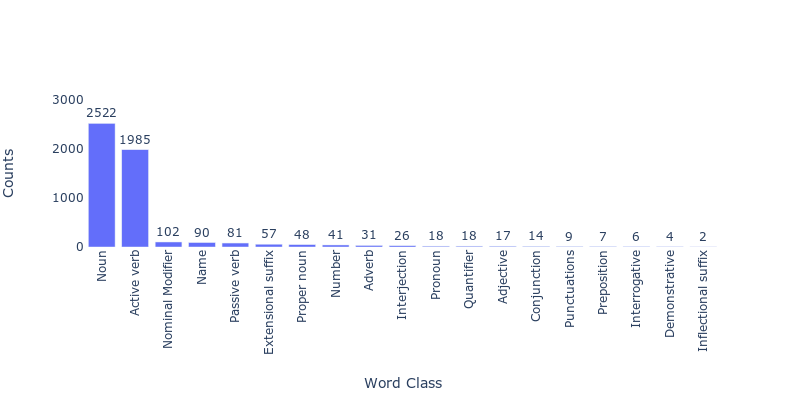}
}
\caption{Distribution of Word Classes.}
\label{fig.5}
\end{center}
\end{figure*}

The IgboAPI dataset encompasses 33 distinct Igbo dialects. Within this dataset, there are 5,095 Igbo words, categorized into various word classes such as nouns, verbs, and adjectives, as shown in Figure \ref{fig.5}. There are 17,979 unique dialectal word variations, complemented by 27,816 example parallel sentences. A `dialectal word' or `dialectal variation of a word' refers to another word having the same meaning as the word of interest but written (and pronounced) in another dialect.
Figure \ref{fig.1} illustrates the distribution of dialectal words across various dialects. The Ngwa dialect stands out with the most extensive collection of dialectal words while the Ikwo dialect has the lowest collection. Figure \ref{fig.2} provides an illustration of the distribution of dialect word counts. We see that many words in the dictionary dataset are multidialectal, highlighting the dialectal complexity prevalent in the Igbo language \cite{10.1145/3146387,dossou2021okwugb}. This intricate nature renders Igbo an intriguing subject for NLP research and underscores the importance of our dataset.

\section{Experiments}
\label{experiments}

To demonstrate the utility of the IgboAPI dataset, we conducted two experiments: one focusing on a semantic lexicon and the other on machine translation. In the following sections, we delve into each of the experiments, covering the rationale, distinctive value of the dialectal IgboAPI dataset, as well as the outcomes and insights.

\subsection{Igbo Semantic Lexicon}
An important use case for the IgboAPI dataset is the creation of an Igbo semantic lexicon. The semantic lexicon is a key resource in developing a semantic tagger for any language. Semantic tagging facilitates the automatic semantic analysis of text which is a key NLP task that lends itself to a variety of applications in natural language understanding and corpus analysis such as information extraction \cite{RaysonEtAl2005}, discourse analysis \cite{Charteris-Black_Seale_2009}, and social media analysis \cite{Charitonidis2017}.

The manual creation of a comprehensive semantic lexicon is a very daunting task of the order of around 1-2 person-years manual research, hence bootstrapping approaches are vital to speed up the development. A commonly used semantic tagset is the UCREL Semantic Analysis System (USAS) \cite{rayson2004ucrel}. We, therefore, leveraged the multilingual structure of the IgboAPI dataset to bootstrap the development of the Igbo semantic lexicon using the existing English semantic tagger via PyMUSAS\footnote{\url{https://ucrel.github.io/pymusas/}}.

\subsubsection{Experimental Methodology}
This outlines our methodology for the ongoing core work and provides a preliminary evaluation of results of the Igbo semantic system developed, at the time of submission. 

There are 5,095 unique entries in the IgboAPI dataset each of which has a given \texttt{wordClass} (attribute 2 in Figure \ref{fig:igboapi-word}). Given the importance of part-of-speech information in semantic tagging, we compared these with the entries in the dictionary created from MasakhaPOS, an Igbo parts-of-speech dataset \cite{dione2023masakhapos}. This process identified 176 co-occurring words that formed the basis of our bootstrapping experiment. These co-occurring words were manually tagged using the USAS semantic tags\footnote{\url{https://ucrel.lancs.ac.uk/usas/USASSemanticTagset.pdf}} hence providing the benchmark for evaluating the coverage of the bootstrapping process.

\subsubsection{Bootstrapping Process}
In creating the Igbo semantic lexicon, the Igbo Words and their English definitions were extracted from the IgboAPI dataset, and the English definitions of each word were labelled using the PyMUSAS RuleBasedTagger pipeline. Counts of the semantic tags for each of the non-function words in the English definitions were sorted from the highest to the lowest. 

The evaluation of the process involved: 1) checking whether the manually assigned semantic tag appeared in the list of automatically transferred tags produced by tagging the English definitions; 2) If so, how highly ranked it is in the list. These scores were indicated with the labels \texttt{Top-X} to \texttt{ALL} where \texttt{X} shows the number of top tags used for the check. For example, \texttt{Top-1} checks if the manually assigned tag is the top tag from the automatic process while \texttt{Top-5} checks if it appears in the top 5 most common semantic tags. \texttt{ALL} records is it appeared at all in the list.

Another aspect of the evaluation was deciding how to represent counts for words with manually assigned multiple semantic tags. In \texttt{Single Count}, each of the multiple tags contributes to the counts individually while \texttt{Average Count} they collectively contribute a total count of 1. Overall the \texttt{Average Count} was a more strict evaluation score.

\subsubsection{Results \& Discussion}
Table \ref{tab:sem_lex_results} shows the scores across different evaluation settings: \texttt{Single Count} and \texttt{Average Count} combined with \texttt{TOP-1}, \texttt{TOP-5}, \texttt{TOP-10}, and \texttt{ALL}.

\begin{table}[!ht]
\begin{center}
\begin{tabularx}{\columnwidth}{ccc}
      \textbf{TOP} & \textbf{Single Count($\%$)} & \textbf{Average Count ($\%$})\\
      \hline
      \noalign{\vskip 2mm} 
      1 & 49.01 & 48.34 \\
      5 & 59.60 & 58.94 \\
       10 & 61.59 & 60.93 \\
      ALL & 64.24 & 63.58 \\
\end{tabularx}
\caption{Single and average coverage for \texttt{TOP-1}, \texttt{TOP-5}, \texttt{TOP-10}, and \texttt{ALL}.}
\label{tab:sem_lex_results}
 \end{center}
\end{table}

The results above clearly show that, with minimal effort, it is perfectly possible to use the IgboAPI resources - dictionary entries and their English definitions - to bootstrap the creation of Igbo Semantic lexicon. It can be observed that even with the strictest evaluation scheme \texttt{TOP-1}, the automatically generated tags correspond with the human-annotated humanly assigned tags about 50\% of the time which is a good indication of its potential.

However, this method is not without limitations and therefore leaves some room for improvement in future work. For example, we could only use the few words (176 words) that appeared in the intersection with the MasakhaPOS dictionary. Also, we could not explore the usefulness of POS tags because the MasakhaPOS and IgboAPI used different Igbo POS tagsets. Another key challenge is reconciling multi-word and sub-word entries in IgboAPI with single full words in MasakhaPOS.

\subsection{Machine translation with the IgboAPI dataset}

\begin{table}[!h]
\resizebox{\columnwidth}{!}{%
\begin{tabular}{p{2cm}p{2cm}p{2cm}p{2cm}}
    &\textbf{Igbo} & \textbf{English} & \textbf{Evaluation}\\
    \hline
    \hline
    \noalign{\vskip 1mm} 
    Standard Igbo & A gwara Amadi \textcolor{blue}{s\`{o}nyere} any\d{i} & Amadi was invited to join us & \cmark \\
    \hline
    \noalign{\vskip 1mm} 

    Dialectal Igbo & A gwara Amadi \textcolor{blue}{s\`{o}nyelu} any\d{i} & Our son-in-law was addressed to Amadi & \xmark \\
  \hline
\noalign{\vskip 3mm} 

  \hline
  \noalign{\vskip 1mm} 

    Standard Igbo & \d{O} b\d{u} ebe ah\d{u} ka ha na-edobe \textcolor{blue}{ngwaagh\={a}} ha & It is there that they are depositing their weapons. & \cmark \\
    \hline
    \noalign{\vskip 1mm} 

     Dialectal Igbo & \d{O} b\d{u} ebe ah\d{u} ka ha na-edobe \textcolor{blue}{ngw\d{o}\d{o}g\d{u}} ha & It is there that they are setting up their solutions. & \xmark \\
    \hline

\end{tabular}
}
\caption{Two examples of the current state-of-the-art Igbo-English MT model giving wrong translations of Igbo sentences once a Standard Igbo word has been substituted with a dialectal variation. }

\label{tab 1}
\end{table}

The existing Igbo-English machine translation models demonstrate limitations in capturing the intricacies of Igbo dialects, as exemplified by a practical illustration in Table \ref{tab 1}, utilizing the current state-of-the-art Igbo-English MT model introduced in \citet{adelani2022few}. This prompts us to consider the pivotal question: how can we improve the existing Igbo machine translation models to attain a deeper understanding of the diverse Igbo dialects? In this section, we demonstrate that through the finetuning of machine translation systems using our multi-dialectal IgboAPI dataset, we can enhance the proficiency of the existing MT models in encoding dialectal Igbo sentences.


\subsubsection{Experimental Methodology}
As we are investigating the ability of MT systems to understand (and therefore encode) Igbo dialects, our primary focus lies on the encoding properties of the MT model. Consequently, our experiment is exclusively centered on the Igbo-English translation direction, and we employ only the latest state-of-the-art MT model for our experiments. 

To achieve our experimental goal, we finetune the model using our repurposed IgboAPI dataset. Subsequently, we assess the translation quality of both the finetuned and non-finetuned versions on our distinct test sets. 
\paragraph{Repurposing the IgboAPI dictionary dataset:}
\label{sec:dialectalInfuse}
From Figure \ref{fig:igboapi-word}, we observe that each Igbo headword has its English translation (in the form of an English definition). We observe the same trend with the example sentences. We also see that for each Igbo word, there are a number of dialectal word variations. Using the Igbo-English words and sentence examples, we created a parallel Igbo-English corpora ($C1$) from the IgboAPI dictionary dataset. $C1$ consists of 143,878 parallel samples: 18,536 parallel samples from the words and 125,342 parallel samples from the example sentences.

In order to augment the corpora with the dialectal representations, we systematically created `dialectally-infused sentences' by replacing each headword in each example sentence with its corresponding dialectal word variations. This expansion led to 135,021 dialectal samples which considerably enriched the dataset, rendering it representative of the diverse linguistic nuances present in various Igbo dialects. Our final dataset, $C2$, used for our MT experiments consists of the corpora from $C1$ as well as the `dialectal sentences'. $C2$ contains a total of parallel 278,899 samples.

\paragraph{The Train and Test Datasets:} For our MT experiment, we created two test sets from $C2$. The first, termed \code{UnseenDialectTestSet}, consists of samples from seven chosen dialects (Afiikpo, Izii, Ezaa, Udi, \d{O}h\d{u}h\d{u}, Ezeagu, and Ogidi). These dialects were chosen due to their linguistic similarities and the few available samples available in these dialects. The \code{UnseenDialectTestSet} test set, which contains 39,126 parallel text, is meant to evaluate the generalization capabilities of the MT model as these dialects were not seen during finetuning. 

The rest of the dataset was randomly split into training, validation, and test sets (80\%:10\%:10\% respectively) as is the standard in machine learning experiments. From this partition, we derived the \code{StandardTestSet} which contains 23,978 parallel text. 



\paragraph{Baseline Model:}
We leveraged the state-of-the-art, multilingual M2M100 \cite{fan2021beyond} translation model, \code{m2m100\_418m\_ibo\_en\_rel\_news}\footnote{\url{https://huggingface.co/masakhane/m2m100_418M_ibo_en_rel_news}} introduced in \citet{adelani2022few,fan2021beyond}. We refer to this model as M2M-IBO-EN in the rest of the paper. 

\paragraph{Training Parameters:} The default parameters from the baseline model were used in fine-tuning the model, except for the epoch which we set to 5. All the experiments were performed on a single GPU (NVIDIA V100). The training lasted for about 10 hours and 5 checkpoints were saved for each epoch.

\paragraph{Evaluation metrics:}
We measured the translation quality using BLEU \cite{papineni2002bleu} and TER \cite{snover2006ter} scores. We included the TER as an extra metric in our evaluation because it provides a deeper insight into the translation errors, helping us understand the extent of discrepancies between the machine-generated translations and the reference translations. The lowest BLEU score attainable is 0 and the highest is 100, indicating perfect translation.

\subsubsection{Ablation study: investigating the effect of dialectal infusion}
\label{ablation}

This ablation study endeavors to examine the influence of the dialectal information in our finetuning dataset on the translation performance. The primary objective is to underscore the significance of the multi-dialectal nature of the IgboAPI dataset.

To perform this study, we constructed two sub-datasets: \code{WithDialect} and \code{WithoutDialect}. To create \code{WithoutDialect}, we extracted 107,723 sentences from $C1$. In so doing, we created a dataset that contains no dialectal variant because the samples here are all from the Standard Igbo headword. For \code{WithDialect}, we randomly sampled 107,723 sentences from the dialectally-infused synthetic samples (Section \ref{sec:dialectalInfuse}). The sub-sampling was done to maintain an equal size of the dataset used for the ablation study. Note that for these two sub-datasets, the words were ignored to ensure that there was no leakage of dialectal information into \code{WithoutDialect}. Finally, the model was separately finetuned on the two sub-datasets, using the same experimental settings, and evaluated on \code{UnseenDialectTestSet}.


\begin{table*}[h]
\begin{tabular}{llll}
\textbf{Igbo}                                                                                & \textbf{English}  & \textbf{\code{WithDialect} prediction} & \textbf{\code{WithoutDialect} prediction} \\
\hline
k\`{u}k\d{o}ta                                             & call together     & gather                            & summarize                          \\
enyaanw\d{\={u}}                                         & the sun; sunlight & the sun                           & grubs                              \\
\`{o}te\`{n}f\d{\={u}} & palm wine tapper  & the palm wine tapper              & he is a kidnapper                 
\end{tabular}
\caption{Examples of translation outputs of the models trained in our ablation studies. We show the Igbo text, reference English text, and the model predictions. With the dialectal information present in the model, it is able to generate a prediction that is closer to the reference text, whereas the non-dialectally infused model mis-translates.}
\label{fig:dialectal-examples}

\end{table*}


\subsection{Result \& Discussion}
Table \ref{tab 2} summarises the results obtained from our finetuning experiments. In the finetuned versions, we observe an improvement of +55.08 BLEU points in StandardTestSet and +51.20 BLEU points in UnseenDialectTestSet. 

\begin{table}[H]
\begin{center}
\begin{tabularx}{\columnwidth}{p{4cm}cc}
      & \textbf{BLEU\big\uparrow} & \textbf{TER\big\downarrow}\\

      \noalign{\vskip 2mm} 

      \textbf{StandardTestSet}&& \\
 \hline
\noalign{\vskip 2mm} 
      M2M-IBO-EN unfinetuned & 16.87 & 66.91 \\
      M2M-IBO-EN finetuned on IgboAPI dataset  & \textbf{71.95} & \textbf{26.71} \\
\noalign{\vskip 2mm} 
      \textbf{UnseenDialectTestSet}&& \\
\hline
\noalign{\vskip 2mm} 
      M2M-IBO-EN unfinetuned & 16.77 & 67.47 \\
    M2M-IBO-EN finetuned on IgboAPI dataset & \textbf{67.91} & \textbf{30.17} \\
 
\end{tabularx}
\caption{The BLEU (higher is better) and TER (lower is better) scores of M2M-IBO-EN on our two test sets: StandardTestSet \& UnseenDialectTestSet.}
\label{tab 2}
 \end{center}
\end{table}


The results of our ablation studies (Section \ref{sec:dialectalInfuse}) are highlighted in Table \ref{fig:dialectal-eval}. One observes that finetuning on the dataset that contains dialectal information leads to a +4.61 improvement in BLEU score. 
\begin{table}[H]
\begin{center}
\begin{tabularx}{\columnwidth}{p{4cm}cc}
      & \textbf{BLEU\big\uparrow}\\

      \noalign{\vskip 2mm}

      \textbf{UnseenDialectTestSet}&& \\
\hline
\noalign{\vskip 2mm} 

            M2M-IBO-EN finetuned on \code{WithoutDialect} & 82.35 \\
    M2M-IBO-EN finetuned on \code{WithDialect}& \textbf{86.96}\\
\end{tabularx}
\caption{Investigating the effect of finetuning on a dataset with dialectal information. We report the BLEU scores of M2M-IBO-EN finetuned with two sub-datasets.}
\label{fig:dialectal-eval}
 \end{center}
\end{table}

While it may seem that the difference in BLEU score is marginal, we show in Table \ref{fig:dialectal-examples} that the translated sentences of these models are heavily affected by the presence (or absence) of the dialectal information during the finetuning process. 

Overall, these findings prove that finetuning our IgboAPI dictionary dataset leads to an enhanced ability of the MT model to comprehend Igbo dialects while translating. Furthermore, the finetuned model performs worse in the UnseenDialectTestSet compared to its performance in the StandardTestSet test set, indicating the difficulty of generalizing to unseen dialects. This underscores the challenge of dialectal machine translation and emphasizes the usefulness of our dataset to this field.

\begin{figure}[h!]
\begin{center}
\resizebox{0.5\textwidth}{!}{
\includegraphics[]{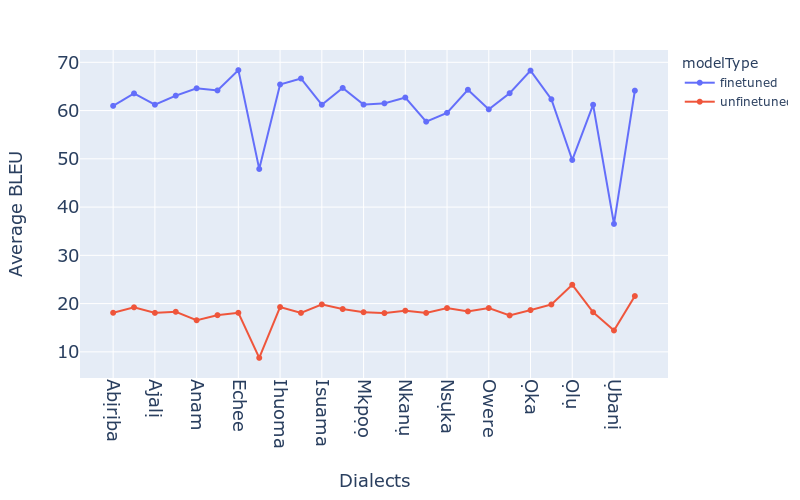}
}
\caption{Average BLEU score for each dialect in the StandardTestSet.}
\label{fig.3}
\end{center}
\end{figure}

\begin{figure}[!ht]
\begin{center}
\resizebox{\columnwidth}{!}{
\includegraphics[]{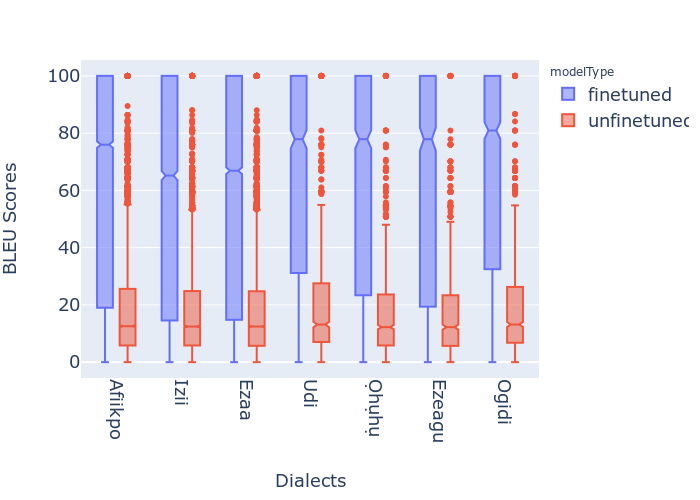}
}
\caption{Boxplot of BLEU scores for each dialect in the UnseenDialectTestSet.}
\label{fig.4}
\end{center}
\end{figure}

In order to offer in-depth analysis into how finetuning on our dataset affects each of the dialects, we provide additional illustrations on the per-dialect BLEU performance for the StandardTestSet in Figure \ref{fig.3} and the UnseenDialectTestSet in Figure \ref{fig.4}.

In both figures, we observe that in each dialect, the finetuned M2M-IBO-EN model consistently outperforms the unfinetuned version. The significant improvement in performance highlights the efficacy of finetuning on our IgboAPI dictionary dataset. In cases where there's a decline in the scores, such as for `Egbema' and `\d{U}bani', this can likely be attributed to the relatively limited training samples available for them compared to other dialects, as indicated in Figure \ref{fig.1}.

Figure \ref{fig.4} shows a boxplot representing the distribution of BLEU scores across all the translation samples within the UnseenDialectTestSet test set for each dialect. A closer examination of the boxplot reveals that the finetuned model's boxplots tend to be higher than those of the unfinetuned model, indicating a general enhancement in the BLEU score distribution with finetuning. However, the substantial spread in the boxplots suggests that the finetuned model does not perform equally well for all sentences in the test set: some samples within the test set exhibit below-average performance.

\section{Conclusion}
In this paper, we present the IgboAPI dataset, a multi-purpose and multi-dialectal Igbo-English dictionary dataset, developed with the aim of enhancing the representation of Igbo dialects. Furthermore, we illustrate the practicality of the IgboAPI dataset through two distinct studies: one focusing on Igbo semantic lexicon and the other on machine translation. In the semantic lexicon project, we successfully established an initial Igbo semantic lexicon for the Igbo semantic tagger by employing a heuristics-based transfer method from English definitions that had been semantically tagged by the PyMUSAS English tagger. In the machine translation study, we demonstrated that by finetuning existing machine translation systems using the IgboAPI dataset, we can significantly improve their ability to encode Igbo sentences with dialectal variations.


\section{Language Resource References}
In this paper, we presented approaches to creating and utilizing a collection of language resources for Igbo. 

\begin{itemize}
    \item We release the IgboAPI Dataset under a CC BY-NC-SA 4.0 DEED license. The dataset can be fully downloaded \href{https://charmed-sycamore-9e5.notion.site/Details-of-the-Dataset-96f3981a329841cd83409eecc4ce1151}{here}.
    \item M2M-IBO-EN Translation Model (\href{https://huggingface.co/masakhane/m2m100_418M_ibo_en_rel_news}{link}).
    \item Igbo Semantic Lexicon (\href{https://github.com/UCREL/Multilingual-USAS}{link}).
    %
\end{itemize}


\bibliographystyle{lrec_natbib}
\bibliography{lrec-coling2024-example}

\begin{thebibliography}{37}
\expandafter\ifx\csname natexlab\endcsname\relax\def\natexlab#1{#1}\fi

\bibitem[{Abbott and Martinus(2018)}]{abbott2018towards}
Jade~Z. Abbott and Laura Martinus. 2018.
\newblock \href {https://arxiv.org/abs/1811.05467v1} {Towards neural machine translation for african languages}.
\newblock \emph{arXiv preprint arXiv: 1811.05467}.

\bibitem[{Abe et~al.(2018)Abe, Matsubayashi, Okazaki, and Inui}]{abe2018multi}
Kaori Abe, Yuichiroh Matsubayashi, Naoaki Okazaki, and Kentaro Inui. 2018.
\newblock Multi-dialect neural machine translation and dialectometry.
\newblock In \emph{Proceedings of the 32nd Pacific Asia Conference on Language, Information and Computation}.

\bibitem[{Adelani et~al.(2022)Adelani, Alabi, Fan, Kreutzer, Shen, Reid, Ruiter, Klakow, Nabende, Chang et~al.}]{adelani2022few}
David~Ifeoluwa Adelani, Jesujoba~Oluwadara Alabi, Angela Fan, Julia Kreutzer, Xiaoyu Shen, Machel Reid, Dana Ruiter, Dietrich Klakow, Peter Nabende, Ernie Chang, et~al. 2022.
\newblock A few thousand translations go a long way! leveraging pre-trained models for african news translation.
\newblock \emph{arXiv preprint arXiv:2205.02022}.

\bibitem[{Agi{\'c} and Vuli{\'c}(2019)}]{agic2019jw300}
{\v{Z}}eljko Agi{\'c} and Ivan Vuli{\'c}. 2019.
\newblock Jw300: A wide-coverage parallel corpus for low-resource languages.
\newblock In \emph{Proceedings of the 57th Annual Meeting of the Association for Computational Linguistics}, pages 3204--3210.

\bibitem[{Almansor and Al-Ani(2017)}]{almansor2017translating}
Ebtesam~H Almansor and Ahmed Al-Ani. 2017.
\newblock Translating dialectal arabic as low resource language using word embedding.
\newblock In \emph{International Conference Recent Advances in Natural Language Processing, RANLP}.

\bibitem[{Anyanwu(2019)}]{anyanwu2019igbo}
Chukwuma Anyanwu. 2019.
\newblock Igbo people in diaspora and the survival of the igbo nation: insights from igbo students association, delta state university, abraka (onye biara ije ga-ala).
\newblock \emph{Journal of African Films and Diaspora Studies}, 2(2):9.

\bibitem[{Anyanwu(2010)}]{anyanwu2010igbo}
O~Anyanwu. 2010.
\newblock Igbo language and its dialects: A challenge for an igbo language teacher.
\newblock \emph{Annals of Modern Education}, 2(1):51--62.

\bibitem[{Asonye(2013)}]{unesco2}
Emmanuel Asonye. 2013.
\newblock Unesco prediction of the igbo language death: Facts and fables.

\bibitem[{Charitonidis et~al.(2017)Charitonidis, Rashid, and Taylor}]{Charitonidis2017}
Christos Charitonidis, Awais Rashid, and Paul~J. Taylor. 2017.
\newblock \href {https://doi.org/10.1007/978-3-319-51049-1_7} {\emph{Predicting Collective Action from Micro-Blog Data}}, pages 141--170. Springer International Publishing, Cham.

\bibitem[{Charteris-Black and Seale(2009)}]{Charteris-Black_Seale_2009}
Jonathan Charteris-Black and Clive Seale. 2009.
\newblock \href {https://doi.org/10.1558/genl.v3i1.81} {Men and emotion talk: Evidence from the experience of illness}.
\newblock \emph{Gender and Language}, 3(1):81–113.

\bibitem[{Dione et~al.(2023)Dione, Adelani, Nabende, Alabi, Sindane, Buzaaba, Muhammad, Emezue, Ogayo, Aremu et~al.}]{dione2023masakhapos}
Cheikh M~Bamba Dione, David Adelani, Peter Nabende, Jesujoba Alabi, Thapelo Sindane, Happy Buzaaba, Shamsuddeen~Hassan Muhammad, Chris~Chinenye Emezue, Perez Ogayo, Anuoluwapo Aremu, et~al. 2023.
\newblock Masakhapos: Part-of-speech tagging for typologically diverse african languages.
\newblock \emph{arXiv preprint arXiv:2305.13989}.

\bibitem[{\d{O}gbal\d{u}(1962)}]{gbal1962kwaOkwuI}
F.~Chidozie \d{O}gbal\d{u}. 1962.
\newblock \href {https://api.semanticscholar.org/CorpusID:160965730} {\d{O}k\d{o}wa-okwu : Igbo-english-english-igbo dictionary}.

\bibitem[{Dossou and Emezue(2021)}]{dossou2021okwugb}
Bonaventure F.~P. Dossou and Chris~C. Emezue. 2021.
\newblock Okwugbé: End-to-end speech recognition for fon and igbo.
\newblock \emph{WINLP}.

\bibitem[{Eberhard et~al.(2020)Eberhard, Simons, and (eds.)}]{ethnologue}
David~M. Eberhard, Gary~F. Simons, and Charles D.~Fennig (eds.). 2020.
\newblock \href {http://www.ethnologue.com} {Ethnologue: Languages of the world. twenty-third edition.}

\bibitem[{Eke(2001)}]{Eke2001}
Julius Eke. 2001.
\newblock \href {https://openlibrary.org/books/OL3779213M/Igbo-English_dictionary} {[link]}.

\bibitem[{Emeka-Nwobia(2019)}]{emekanwobia19language}
Ngozi~Ugo Emeka-Nwobia. 2019.
\newblock \href {https://doi.org/10.1007/978-3-030-02438-3_33} {\emph{Language Endangerment in Nigeria: The Resilience of Igbo Language}}, pages 1643--1655. Springer International Publishing.

\bibitem[{Ezeani et~al.(2020)Ezeani, Rayson, Onyenwe, Uchechukwu, and Hepple}]{ezeani2020igbo}
Ignatius Ezeani, Paul Rayson, Ikechukwu Onyenwe, Chinedu Uchechukwu, and Mark Hepple. 2020.
\newblock Igbo-english machine translation: An evaluation benchmark.
\newblock \emph{arXiv preprint arXiv:2004.00648}.

\bibitem[{Fan et~al.(2021)Fan, Bhosale, Schwenk, Ma, El-Kishky, Goyal, Baines, Celebi, Wenzek, Chaudhary et~al.}]{fan2021beyond}
Angela Fan, Shruti Bhosale, Holger Schwenk, Zhiyi Ma, Ahmed El-Kishky, Siddharth Goyal, Mandeep Baines, Onur Celebi, Guillaume Wenzek, Vishrav Chaudhary, et~al. 2021.
\newblock Beyond english-centric multilingual machine translation.
\newblock \emph{The Journal of Machine Learning Research}, 22(1):4839--4886.

\bibitem[{George(1995)}]{wordnet}
Miller George. 1995.
\newblock \href {https://dl.acm.org/doi/10.1145/219717.219748} {Word net}.
\newblock \emph{Communications of the ACM}.

\bibitem[{Green(1971)}]{green1971}
Margaret~M. Green. 1971.
\newblock \href {https://doi.org/10.2307/1159442} {Igbo: A. learner's dictionary. by beatrice f. welmers and william e. welmers. (a research project financed jointly by the university of california, los angeles, and the united states peace corps.) ucla: African studies center, 1968. pp. x 397. no price given.}
\newblock \emph{Africa}, 41(2):180–181.

\bibitem[{Hirschberg and Manning(2015)}]{hirschberg2015advances}
Julia Hirschberg and Christopher~D Manning. 2015.
\newblock Advances in natural language processing.
\newblock \emph{Science}, 349(6245):261--266.

\bibitem[{Igboanusi(2017)}]{igboanusiglossary}
Herbert Igboanusi. 2017.
\newblock \emph{English-Igbo Glossary of HIV, AIDS and Ebola-related terms}.

\bibitem[{IgboAPI()}]{dictEditingStandards}
IgboAPI.
\newblock Igboapi: Dictionary editing standards.
\newblock \url{https://bit.ly/3tH06UT}.
\newblock [Accessed 20-10-2023].

\bibitem[{Joshi et~al.(2020)Joshi, Santy, Budhiraja, Bali, and Choudhury}]{joshi-etal-2020-state}
Pratik Joshi, Sebastin Santy, Amar Budhiraja, Kalika Bali, and Monojit Choudhury. 2020.
\newblock \href {https://doi.org/10.18653/v1/2020.acl-main.560} {The state and fate of linguistic diversity and inclusion in the {NLP} world}.
\newblock In \emph{Proceedings of the 58th Annual Meeting of the Association for Computational Linguistics}, pages 6282--6293, Online. Association for Computational Linguistics.

\bibitem[{Mbah(2021)}]{Mbah2021}
B.M~Onyemaechi Mbah. 2021.
\newblock \href {https://koha.nou.edu.ng/cgi-bin/koha/opac-detail.pl?biblionumber=11322#} {Igbo dictionary osanye okwu igbo na nkowa ya}.

\bibitem[{Mbonu et~al.(2022)Mbonu, Chukwuneke, Paul, Ezeani, and Onyenwe}]{mbonu2022igbosum1500}
Chinedu Mbonu, Chiamaka Chukwuneke, Roseline Paul, Ignatius Ezeani, and Ikechukwu Onyenwe. 2022.
\newblock Igbosum1500-introducing the igbo text summarization dataset.
\newblock In \emph{3rd Workshop on African Natural Language Processing}.

\bibitem[{Nekoto et~al.(2020)Nekoto, Marivate, Matsila, Fasubaa, Fagbohungbe, Akinola, Muhammad, Kabongo~Kabenamualu, Osei, Sackey, Niyongabo, Macharm, Ogayo, Ahia, Berhe, Adeyemi, Mokgesi-Selinga, Okegbemi, Martinus, Tajudeen, Degila, Ogueji, Siminyu, Kreutzer, Webster, Ali, Abbott, Orife, Ezeani, Dangana, Kamper, Elsahar, Duru, Kioko, Espoir, van Biljon, Whitenack, Onyefuluchi, Emezue, Dossou, Sibanda, Bassey, Olabiyi, Ramkilowan, {\"O}ktem, Akinfaderin, and Bashir}]{nekoto-etal-2020-participatory}
Wilhelmina Nekoto, Vukosi Marivate, Tshinondiwa Matsila, Timi Fasubaa, Taiwo Fagbohungbe, Solomon~Oluwole Akinola, Shamsuddeen Muhammad, Salomon Kabongo~Kabenamualu, Salomey Osei, Freshia Sackey, Rubungo~Andre Niyongabo, Ricky Macharm, Perez Ogayo, Orevaoghene Ahia, Musie~Meressa Berhe, Mofetoluwa Adeyemi, Masabata Mokgesi-Selinga, Lawrence Okegbemi, Laura Martinus, Kolawole Tajudeen, Kevin Degila, Kelechi Ogueji, Kathleen Siminyu, Julia Kreutzer, Jason Webster, Jamiil~Toure Ali, Jade Abbott, Iroro Orife, Ignatius Ezeani, Idris~Abdulkadir Dangana, Herman Kamper, Hady Elsahar, Goodness Duru, Ghollah Kioko, Murhabazi Espoir, Elan van Biljon, Daniel Whitenack, Christopher Onyefuluchi, Chris~Chinenye Emezue, Bonaventure F.~P. Dossou, Blessing Sibanda, Blessing Bassey, Ayodele Olabiyi, Arshath Ramkilowan, Alp {\"O}ktem, Adewale Akinfaderin, and Abdallah Bashir. 2020.
\newblock \href {https://doi.org/10.18653/v1/2020.findings-emnlp.195} {Participatory research for low-resourced machine translation: A case study in {A}frican languages}.
\newblock In \emph{Findings of the Association for Computational Linguistics: EMNLP 2020}, pages 2144--2160, Online. Association for Computational Linguistics.

\bibitem[{Nnaji(1985)}]{Nnaji1985}
H.~I. Nnaji. 1985.
\newblock \href {https://openlibrary.org/works/OL4901875W/Modern_English-Igbo_dictionary?edition=key%3A%2Fbooks%2FOL17303699M} {Modern english-igbo dictionary by h. i. nnaji}.

\bibitem[{Nwankwere et~al.(2017)Nwankwere, Mmadike, and Eme}]{nwankwere2017safeguarding}
Angela~\d{U}N Nwankwere, Benjamin~I Mmadike, and Cecilia~A Eme. 2017.
\newblock Safeguarding the igbo language through teaching igbo children in diaspora.
\newblock \emph{Ogirisi: a new journal of African studies}, 13:166--186.

\bibitem[{Nwaozuzu(2008)}]{nwaozuzu2008dialects}
G.I. Nwaozuzu. 2008.
\newblock \href {https://books.google.de/books?id=GodjYgEACAAJ} {\emph{Dialects of the Igbo Language}}.
\newblock University of Nigeria Press.

\bibitem[{Onyemelukwe(2019)}]{article}
Ifeoma Onyemelukwe. 2019.
\newblock Language endangerment: The case of the igbo language.
\newblock \emph{Impact}, 7:213--224.

\bibitem[{Onyenwe et~al.(2018)Onyenwe, Hepple, Chinedu, and Ezeani}]{10.1145/3146387}
Ikechukwu~E Onyenwe, Mark Hepple, Uchechukwu Chinedu, and Ignatius Ezeani. 2018.
\newblock \href {https://doi.org/10.1145/3146387} {A basic language resource kit implementation for the igbonlp project}.
\newblock \emph{ACM Trans. Asian Low-Resour. Lang. Inf. Process.}, 17(2).

\bibitem[{Opara(2016)}]{oparaefefct2016}
Chioma~Daberechukwu Opara. 2016.
\newblock The efefct of modern technology on the igbo traditional system of communication.
\newblock \emph{ISSN 1119-961 X}, page~75.

\bibitem[{Papineni et~al.(2002)Papineni, Roukos, Ward, and Zhu}]{papineni2002bleu}
Kishore Papineni, Salim Roukos, Todd Ward, and Wei-Jing Zhu. 2002.
\newblock Bleu: a method for automatic evaluation of machine translation.
\newblock In \emph{Proceedings of the 40th annual meeting of the Association for Computational Linguistics}, pages 311--318.

\bibitem[{Rayson et~al.(2005)Rayson, Cosh, and Sawyer}]{RaysonEtAl2005}
P.~Rayson, K.~Cosh, and P.~Sawyer. 2005.
\newblock \href {https://doi.org/10.1109/TSE.2005.129} {Shallow knowledge as an aid to deep understanding in early phase requirements engineering}.
\newblock \emph{IEEE Transactions on Software Engineering}, 31(11):969--981.

\bibitem[{Rayson et~al.(2004)Rayson, Archer, Piao, and McEnery}]{rayson2004ucrel}
Paul Rayson, Dawn~E Archer, Scott~L Piao, and Tony McEnery. 2004.
\newblock The ucrel semantic analysis system.
\newblock In \emph{Proceedings of the workshop on Beyond Named Entity Recognition Semantic labelling for NLP tasks, in association with LREC-04}, pages 7--12. European Language Resources Association.

\bibitem[{Snover et~al.(2006)Snover, Dorr, Schwartz, Micciulla, and Makhoul}]{snover2006ter}
Matthew Snover, Bonnie Dorr, Richard Schwartz, Linnea Micciulla, and John Makhoul. 2006.
\newblock A study of translation edit rate with targeted human annotation.
\newblock In \emph{Proceedings of the 7th Conference of the Association for Machine Translation in the Americas: Technical Papers}, pages 223--231.

\end{thebibliography}


\label{lr:ref}
\bibliographystylelanguageresource{lrec-coling2024-natbib}
\bibliographylanguageresource{languageresource}




\end{document}